\documentclass[acmsmall]{acmart}

\usepackage{mathastext}

\usepackage{subcaption}
\usepackage{footmisc}

\AtBeginDocument{%
  \providecommand\BibTeX{{%
    \normalfont B\kern-0.5em{\scshape i\kern-0.25em b}\kern-0.8em\TeX}}}

\acmJournal{TALLIP}
\acmVolume{19}

\begin{document}
\title{BERT Based Multilingual Machine Comprehension in English and Hindi}


\author{Somil Gupta}
\email{somilgupta@umass.edu}
\orcid{0000-0002-6833-9989}

\author{Nilesh Khade}
\email{nkhade@umass.edu}
\orcid{0000-0002-5877-571X}
\affiliation{%
  \institution{University of Massachusetts Amherst}
  \city{Amherst}
    \state{MA}
    \postcode{01003}
  \country{United States}}


\begin{abstract}
Multilingual Machine Comprehension (MMC) is a Question-Answering (QA) sub-task that involves quoting the answer for a question from a given snippet, where the question and the snippet can be in different languages. Recently released multilingual variant of BERT (m-BERT), pre-trained with 104 languages, has performed well in both zero-shot and fine-tuned settings for multilingual tasks; however, it has not been used for English-Hindi MMC yet. We, therefore, present in this article, our experiments with m-BERT for MMC in zero-shot, mono-lingual (e.g. Hindi Question-Hindi Snippet) and cross-lingual (e.g. English Question-Hindi Snippet) fine-tune setups. These model variants are evaluated on all possible multilingual settings and results are compared against the current state-of-the-art sequential QA system for these languages. Experiments show that m-BERT, with fine-tuning, improves performance on all evaluation settings across both the datasets used by the prior model, therefore establishing m-BERT based MMC as the new state-of-the-art for English and Hindi. We also publish our results on an extended version of the recently released XQuAD dataset, which we propose to use as the evaluation benchmark for future research.
\end{abstract}



\maketitle

\section{Introduction}

\par The Task of Question Answering (QA) has been represented as an Information Retrieval (IR) task, in which questions asked in natural language by humans can be correctly answered automatically using structured databases, natural language documents or web pages. It is a lot different from a standard search engine approach as search engines output the relevant documents for a query while QA systems output the relevant information across those documents. Two major factors influence the quality of output of a QA system, first, the spectrum of information resource used to seek the result and, second, the understanding of that information to be able to output the requisite answer effectively. In the current age of information explosion, extracting exact information, even for a simple query, requires heavy resources for computation and evaluation. Therefore, most research in this domain focuses on the second factor. This factor of understanding the inherent contextual information is termed as \textbf{Machine Reading Comprehension (MRC)} in recent NLP literature. 
\par The challenge of MRC is addressed by conversational AI systems in which QA agents need to provide concise, direct answers to user queries through conversation, extracted from a text given as context. This limits the focus on natural language understanding in a resource bounded manner and aids evaluation by restricting the scope of possible answers.

\par An important challenge faced by QA systems is the presence of knowledge base in different languages, thus restricting the source of monolingual systems. This can be efficiently handled with multilingual systems which can comprehend the syntactic and semantic structure of multiple languages simultaneously. These systems are known as \textbf{Multilingual Question Answering (MQA)} systems and are able to comprehend question in one language and answer using resources in another. MQA is a much needed requirement in IR, especially for non-Latin languages as most of the knowledge on the web is present is English, making search in foreign languages like Hindi, Japanese and Mandarin highly restricted. \textbf{Multilingual Machine Comprehension(MMC)} can then be called as a special case of MQA, where context is given as an excerpt from the available text and answer is required as a span of this text. 

\par Various approaches have been employed to solve the MMC challenge across different language pairs, however, the recent multilingual variant of BERT \cite{bert} called m-BERT -- a single, deep contextualized language model pre-trained from monolingual corpora in 104 languages -- has given state-of-art-results across many language pairs like English-French and English-Japanese \cite{multilingualQA}. Surprisingly, m-BERT also performs well at zero-shot cross-lingual model transfer, in which task-specific annotations in one language are used to fine-tune the model for evaluation in another \cite{bert-multilingual}. This encourages us to use m-BERT for English-Hindi MMC where it has not been tried for MQA setting yet. The latest state-of-the-art \cite{latestMQA} uses a sequential model with joint training on English and Hindi to solve the problem. 
\par One of the major drawbacks for research in English-Hindi MMC currently is the lack of a standard evaluation dataset. The prior state-of-the-art \cite{latestMQA} used a translated subsection of the SQuAD \cite{squad2.0} dataset for evaluation, however, it is not publicly available and requires significant preprocessing as the instances are not in SQuAD-format and the translated answers have a lot of inconsistencies due to them being machine-translated. Though we managed to retrieve the dataset from the authors and preprocess it for comparison against the reported results, we do not recommend it for future research. DeepMind recently released  XQuAD \cite{xquad} as a multilingual evaluation benchmark, which comprises a subsection of the SQuAD v1.1 \cite{squad} development set translated into ten languages (including Hindi) by professional translators. We further extend this dataset by creating cross-lingual variants (e.g. English Question and Hindi Answer) and publish our results on this extended XQuAD dataset, which we propose as the new evaluation benchmark for future endeavours in this domain. 
\par  As part of our experiments, we use the preprocessed datasets of the previous model \cite{latestMQA} as SQuAD \cite{squad} style synthetic datasets, along with extended XQuAD dataset. We fine-tune m-BERT for MMC under \textit{zero-shot} (no prior Hindi fine-tuning), Hindi \textit{monolingual} and \textit{cross-lingual} augmentation settings. Finally, all the model settings are evaluated for all possible multilingual use-cases (cross- and mono-) over the datasets.  Although the zero-shot setting does not beat the benchmark, mono- and cross-lingual fine-tune settings show significant improvement over the baseline \cite{latestMQA}. Major contributions of our work are therefore \textit{\textbf{(i) }establishing m-BERT based MMC as the new state-of-the-art for English and Hindi, and, \textbf{(ii) } standardizing evaluation in English-Hindi MMC by proposing our extended version of XQuAD as the new benchmark.}

\section{Related Work}
\label{related_work}
\par Although MQA has had its own track at the CLEF\footnote{\url{http://www.clef-initiative.eu/home}} since 2003 \cite{multilingual-trec}, much of the work in MQA has been centered around using Machine Translation to bring either of the two in the other language, followed by application of QA techniques \cite{translate}. However, with the introduction of m-BERT \footnote{\label{mbert}\url{https://github.com/google-research/bert/blob/master/multilingual.md}} \cite{bert}, recent end-to-end neural QA systems have emerged \cite{multilingualQA} which have performed well even in zero-shot setting \cite{bert-multilingual}. 

\par However, most of the research in the field of QA systems has happened in resource-rich languages such as English, which is difficult to port into other relatively low-resource languages. In the NLP literature on indic \footnote{\url{https://en.wikipedia.org/wiki/Indo-Aryan\_languages}} languages, we see very few attempts in the direction of MQA. The first hindi monolingual QA system \cite{prashnottar} was introduced in 2012, however, it was as recent as 2018 when the first MQA system for Hindi and English was released \cite{mmqa}. The system followed an IR-based approach of similarity computation and ranking for answering questions in either language, using comparable English and Hindi documents. The questions and documents formed their benchmark dataset called multi-domain multilingual QA (MMQA). 

\par The current state-of-the-art English-Hindi MQA \cite{latestMQA} by the same authors is perhaps the first MMC system in this domain. It applies a deep learning approach using Sequence models(GRU). Their approach relies on jointly training the model on mapped context-question pairs of both languages.
First, it involves generation of a context snippet from given MMQA resources, using a graph-based language-independent algorithm which leverages the lexico-semantic similarity between the sentences. Then, it tries to learn the question representation by soft aligning words in both Hindi and English questions. Finally, the question-aware snippet representations are passed to an answer extraction layer, which extracts answer span from both the snippets simultaneously. 
\par While both approaches are able to handle multilingual inputs without having to
translate them into a single language, our approach differs significantly from \cite{latestMQA}. Their approach relies on sequential modelling using bi-GRU \cite{bi-GRU} which has become lack-lustre since BERT gained popularity as the state-of-the-art language model for MRC task. Our approach, on the other hand, leverages the multilingual contextual reasoning of m-BERT for MMC. \textit{None of the recent literature has used m-BERT to solve the MMC problem for Hindi-English pair (to the best of our knowledge)}. However, given the recent success of m-BERT in MQA in other languages \cite{liu-etal-2019-xqa}, our aim seems like a good proposition.

\section{Background}
\par Our approach is mainly built on top of m-BERT whose architecture is the same as that of monolingual BERT \cite{bert}, except for the corpus used in pre-training. Therefore, it is important to discuss MRC based fine-tune architecture of the original BERT and the reasons behind choosing m-BERT before discussing our proposed approach.  
\subsection{Fine-tuning BERT for Machine Comprehension Task}
\begin{figure}[h]
\includegraphics[width=6cm]{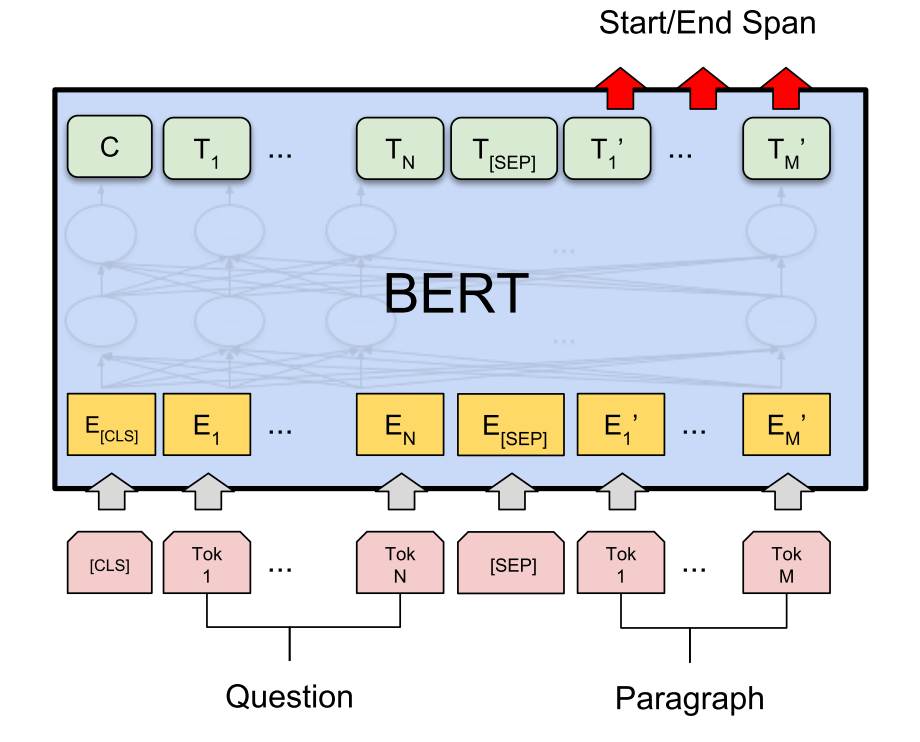}
 \caption{Machine comprehension based fine-tuning procedure for BERT. Apart from output layers, the same architecture can be used by different downstream tasks, initializing all layers with the same pretrained model parameters. During fine-tuning, all parameters are fine-tuned. [CLS] is a special symbol added in front of every input example whose corresponding output embedding represents contextualized embedding for the entire sequence, while [SEP] is a special separator token (e.g. separating questions/answers). Each output embedding represents the contextualized embedding of corresponding input token which is passage and question-aware. \textit{Source: \cite{bert}}}.
 \label{fig:bert_ft}
\end{figure}
\par In the MRC based fine-tuning task as shown in figure \ref{fig:bert_ft}, input question and passage are tokenized and presented as a single packed sequence. Embedding $E_i$ and $E'_i$ stand for the question and answer tokens respectively, while $T_i$ and $T'_i$ denote their corresponding contextualized output embeddings respectively. Input embeddings incorporate the lexical embeddings, positional embeddings(to distinguish position) and segment embeddings (to differentiate question from context tokens). BERT performs contextual integration such that each output embedding is question-aware and context-aware. Additional component introduced during fine-tuning is the output classification layer characterized by the start vector $S \in \mathbf{R}^H$ and an end vector $E \in \mathbf{R}^H$. 
\par Answer span is decided by calculating the softmax on all the passage output embeddings to detect the start word index and end word index of the answer from the given paragraph. So, the probability of a word $T'_i$ to be the start index is $P_i = \frac{e^{S.T'_i}}{\sum_j e^{S.T'_j}}$. The similar formula is used to detect the most probable end of the answer span. The net score for every possible span from position $i$ to position $j$ is denoted as $S.T'_i + E.T'_j$ . So, the final prediction for the answer is the span for which this combined score is maximum. For the training, summation of log-likelihood of the correct start and end positions is used as an objective function.
\par Although the above MRC approach provides contiguous span, it can be used for non-contiguous spans by finding the span that maximizes the F1 score. 

\subsection{Why m-BERT as a Base Model?}

\par M-BERT is a multilingual variant of BERT, with exactly the same architecture and APIs. Both multilingual and monolingual language model variants are pretrained, in an unsupervised manner, using the same Masked Language Modelling(MLM) and Natural Language Inference(NLI) approaches outlined in \cite{bert}. However, while the original monolingual BERT was trained in English corpus of Wikipedia and BooksCorpus, m-BERT is trained on a concatenation of mono-lingual Wikipedia corpora from 104 languages, thus possessing multilingual text understanding.
\par It has been researched in \citep{bert-multilingual} that m-BERT  enables a very straightforward approach to zero-shot cross-lingual model transfer. It is easier to fine-tune the model for a specific task in any language (known by the model) with the help of data from a readily available language, and evaluating that task in another desired, but less-resourced language. This helps to generate models which generalize information across languages. 
\par There might be some apprehension with zero-shot cross lingual transfer between English and Hindi as both have different scripts (Latin and Devanagiri respectively). However, it has been empirically shown that m-BERT is also able to transfer between languages written in different scripts \cite{multilingualQA}, even though they may have zero lexical overlap, thus indicating that m-BERT captures multilingual representations \cite{cross-lingual-effective-mbert}. Therefore, we have leveraged the fact that m-BERT’s multilingual representation is able to map learned structures onto new vocabularies and can work well in English-Hindi setting. 

\section{Proposed Approach}
\label{proposed_approach}
\par We train m-BERT on MMC task for English and Hindi in both zero-shot and fine-tuned settings. These fine-tuned models are then evaluated in cross-lingual and mono-lingual fashion.  
\subsection{Fine-tune Settings}
\label{finetune_settings}
\par These settings describe the model state before evaluation i.e. how the model was fine-tuned.
\begin{itemize}
    \item \textbf{Zero Shot baseline} i.e. fine-tuning the model for MMC on a language different from the one in which it is evaluated. For this, the model is fine-tuned on the SQuAD dataset \cite{squad} (in English) and then evaluated on instances with Hindi context(passage). This helps to test whether m-BERT performs well without any prior fine-tuning in Hindi. 
    
    \item \textbf{Hindi mono-lingual QA augmentation:} In this approach, we augment the zero-shot MMC model from the first approach, with the Hindi QA training set (mono-lingual QA with both question and answer in Hindi). Although this ensures the model is trained in Hindi reading comprehension, it does not explicitly establish any cross-lingual relationship between an English QA instance and its corresponding Hindi counterpart. 
    
    \item \textbf{Hindi cross-lingual QA augmentation}: This setting is a continuation to the second approach that involves further augmenting the fine-tuned model with a cross-lingual training set, for example, the set with questions in English and answers in Hindi. This may help achieve cross-lingual relationship between the languages. The cross-lingual variant to train is explored in the experiment section. 
\end{itemize}
\par Aiming on different attributes of m-BERT in each fine-tune setting can also help identify which aspect is already handled by the m-BERT intrinsically, for instance, model may be capable of capturing cross-lingual ability between the two languages by itself, in which case, the second and the third settings would not have substantial difference in their cross-lingual evaluation.
\par As an additional approach, we had also thought of fine-tuning on joint input of English and Hindi QA similar to \cite{latestMQA}. The approach was aimed at explicitly instilling translational ability in the model for both the languages, as both the text and its translation would be input simultaneously. However, as we would see in the results, our proposed training flow achieved state-of-the-art performance, indicating that m-BERT can handle multilingual QA without explicit training in translation. 

\subsection{Evaluation Settings}
\label{eval_settings}
\par In order to fully evaluate an MMC model, we need to evaluate the model on all possible multilingual settings (cross and mono) using English and Hindi. Following are the different evaluation settings. These settings are similar to the ones used in \cite{latestMQA}, expect for the ones which require joint answering in both languages. 
\begin{itemize}
\item $Q_E$ - $P_E$: Both question and passage are in English. This setting tests English monolingual comprehension ability of the model. 
\item $Q_H$ - $P_H$: Both question and passage are in Hindi. This setting tests Hindi monolingual comprehension ability of the model. 
\item $Q_E$ - $P_H$: The question is in English and the answer exists in the Hindi passage. This setting tests cross-lingual comprehension capability of the model.
\item $Q_H$ - $P_E$: The question is in Hindi and the answer exists in the English snippet. This setting too tests the cross-lingual comprehension capability of the model.
\end{itemize}
\par For metrics, we would rely on the Exact Match (EM) and F1 score. EM measures the percentage of predictions that match any one of the ground truth answers exactly. F1 score is a looser metric that measures the average overlap between the prediction and ground truth answer.

\section{Data}
\label{datasets}
\par \cite{latestMQA} evaluates its model on a translated subsection of the SQuAD dataset and the MMQA dataset \cite{mmqa}. Since it is our baseline model, we use the same datasets for our evaluation as well. Besides, we also evaluate our model on the XQuAD dataset as benchmark for future research. SQuAD and its translated version are also used for fine-tuning. 
\par M-BERT has predefined functionality for MRC on SQuAD dataset. To reuse it, we needed to transform the datasets to SQuAD format as outlined in Figure \ref{fig:squad}. Besides, our evaluation approach defined in section \ref{eval_settings} requires us to create test variants for all different multilingual settings. Fine-tuning too needs $Q_H-P_H$ and $Q_E-P_H$ variants of the training set. Therefore, the datasets need preprocessing as well.  

\subsection{SQuAD}
\label{squad_sec}
\begin{figure}[h]
\includegraphics[scale=0.4]{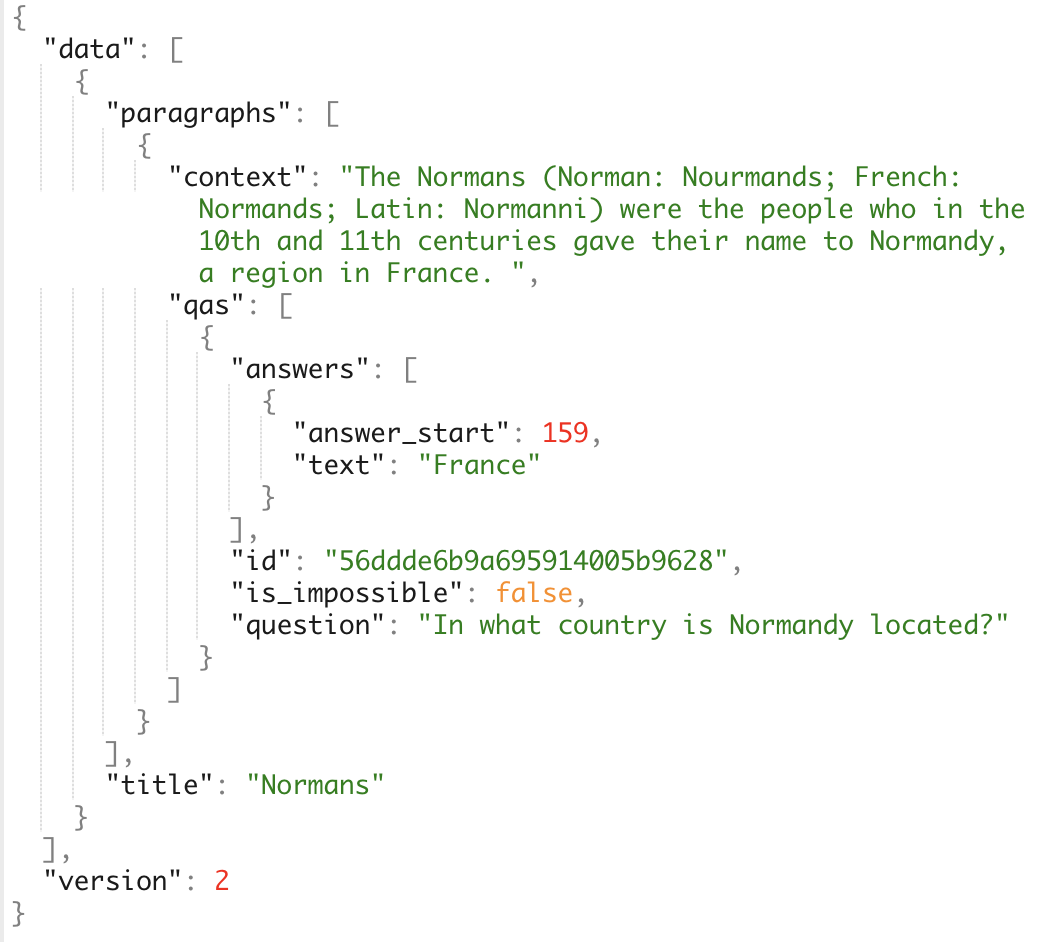}
 \caption{An example of the SQuAD dataset. Each dataset contains an array of instances, each belonging to a Wikipedia article with title and paragraphs. Each paragraph JSON consists of an excerpt and a set of question answers pertaining to that context. Each question has a unique id, answer(text with start index). An unanswerable question  is marked as impossible. }
 \label{fig:squad}
\end{figure}

\textit{Standford Question Answering Dataset (SQuAD)} is a monolingual, single-turn MRC dataset released by the Stanford NLP group consisting of questions posed by crowd workers on a set of Wikipedia articles. It is the first large-scale, high quality dataset released  in  MRC  that  greatly  ushered  research in Machine Comprehension.  Version 1.0 \cite{squad} consists of 100K+ questions from a variety of answer types while version 2.0 \cite{squad2.0} lays emphasis on unanswerable questions with over 50K additional questions. Figure \ref{fig:squad} provides the format of SQuAD 2.0.

\par The authors of \cite{latestMQA} created a translated SQuAD subset 
containing tuples of the format (question, passage,
answer, start token index, end token index) for both English and Hindi. These tuples are generated using randomly chosen 18,454 SQuAD instances and their corresponding Hindi machine-translation (using Google translator). The dataset has a \textit{train-validation-test split of  10,454 to 2K to 6K} respectively.  Since the site \footnote{\url{http://www.iitp.ac.in/~ai-nlp-ml/resources.html}} stated in their paper did not contain the dataset, we had to fetch it directly  from the authors. 
\subsubsection{Preprocessing}: The translated SQuAD dataset had numerous inconsistencies which made creation of variants difficult, for example, the given answer text had extra or missing characters which prevented substring-match with the given context, answer start indexes did not correspond to exact location of answer text, and there were no question IDs or titles to tie to original text. To mitigate these issues, the text was sanitized using pattern match-and-replace style sanity checks (e.g. removing space before (,), adding dots in abbreviations, etc.), correct index of exact start index of the answer text was located using substring find operation, and questions in original SQuAD 2.0 dataset were looked up to locate question IDs and regroup disjoint questions using context passage as pivot. Training split helped create fine-tuning variants, while test split helped create evaluation variants. 

\subsection{MMQA}
\label{mmqa_sec}
Multi-domain Multilingual QA (MMQA) dataset\footnote{https://github.com/deepaknlp/MMQA} was released by \cite{mmqa} as the first benchmark dataset for English-Hindi MQA. It consists of around 5,500 QA pairs formulated by human annotators over 500 articles curated from 6 different domains, with comparable English and Hindi document for each article. Questions may be factoid or short descriptive type. Answers can be abstractive (may not be exact quotations from the text, or contiguous spans of text), however,  due to majority of questions being factoid, answers are generally contiguous spans from articles. 
\par \cite{latestMQA} use a subset of about 3960 QA pairs from MMQA for evaluation and generate snippet for the pairs using their novel algorithm. We evaluate our model on this subset alongwith the generated snippets and compare our results against the ones reported in their paper. 
\subsubsection{Preprocessing:} The given set of QA pairs with their snippets additionally contains a field identifying the instance as English only, Hindi only or both. We use this field to bucket the instances into mono and cross lingual variants. While transforming the instances to SQuAD format, we only populate the answer text and not the start index, as the answers are not exact substrings. The SQuAD evaluation script does not use start index for calculating F1 and EM scores. 

\subsection{XQuAD}
\label{xquadsec}
\par XQuAD (Cross-lingual Question Answering Dataset)\footnote{\label{xquad_url}\url{https://github.com/deepmind/xquad}} is a benchmark dataset published by DeepMind for evaluating multilingual QA performance. Released as part of \cite{xquad}, the dataset consists of monolingual versions of a subset of 240 paragraphs and 1190 QA pairs from the development set of SQuAD v1.1 \cite{squad} translated into ten languages including Hindi. It is the most appropriate benchmark evaluation set for English-Hindi MMC, because:
\begin{enumerate}
    \item The translations are accomplished by human translators, making the dataset more reliable for evaluation than the previous two datasets.
    \item Being a translated version of SQuAD, the dataset is compliant to MRC task API of BERT, without any additional preprocessing.
    \item Universal acceptance of SQuAD as default MRC benchmark implies that XQuAD can also be considered the de-facto benchmark dataset for MMC.
\end{enumerate}  
\subsubsection{Extension:} The dataset only contains monolingual variants of the SQuAD subset, however for complete evaluation of multilingual ability, it is important to test on cross-lingual variants as well, where the question and the context are in different languages. We create these cross-lingual variants using the monolingual variants of English and Hindi which are already present.

\section{Experiment}
\par The experiment was conducted for all the approaches outlined in section \ref{proposed_approach} and SQuAD-style variants of the datasets were created by preprocessing as mentioned in section \ref{datasets}. 
\subsubsection{Training phase} For each training setup, 10 parallel runs of m-BERT fine-tuning were conducted for 30,000 steps with hyper-parameters as defined in table \ref{tab:mbert}. 
\begin{itemize}
    \item \textbf{Zero-shot:}, \textit{BERT-Base, Multilingual Cased model\footref{mbert} (104 languages, 12-layer, 768-hidden, 12-heads, 110M parameters)} was fine-tuned on the entire training set of original SQuAD 2.0 dataset, containing 150K QA pairs. No fine-tuning on Hindi was involved in this setup. 
    \item \textbf{Hindi monolingual QA augmentation}: Each zero-shot checkpoint was further fine-tuned on $Q_H-P_H$ variant of the training split of translated SQuAD dataset, which consists of around 10.5K QA pairs. There was a one-to-one mapping between a zero-shot checkpoint and the new checkpoint, thus maintaining the same number of checkpoints for this setup.  
    \item \textbf{Hindi cross-lingual QA augmentation}: Each checkpoint after Hindi monolingual QA augmentation was further fine-tuned for this setup on $Q_E-P_H$ variant of the training split of translated SQuAD dataset. Similar to the previous setup, one-to-one mapping was maintained between the checkpoints. It is important to note that the choice of cross-lingual variant for augmentation ($Q_E-P_H$ or $Q_H-P_E$) was completely arbitrary and any of them would have given better results as we see in the later section. 
\end{itemize}

\begin{table}[h!]
\caption{Important hyperparameter values for m-BERT fine-tuning and evaluation.}
\begin{tabular}{l|l}
 \hline
Hyperparameter           &  Value \\ \hline
Lower the case for input             & False          \\ 
Max (sequence, query, answer) length & 384, 64, 30    \\ 
Train, predict batch sizes           & 12,8           \\ 
Learning rate, warmup proportion     & 5e-5, 0.1      \\ 
Number of epochs                     & 3.0            \\ \hline
\end{tabular}%
\label{tab:mbert}
\end{table}

\subsubsection{Evaluation phase} 
\label{evaluation}
For each training setup, all the checkpoints were evaluated on every multilingual evaluation variant of every dataset, thus creating a total of 10 checkpoints $\times$ 4 variants  $\times$ 3 datasets $ = 120$ prediction sets for each training setup. SQuAD's evaluation script\footnote{\url{https://rajpurkar.github.io/SQuAD-explorer/}} was used for generating F1 and EM scores on each such prediction set. Finally, averaged scores across all checkpoints for each dataset, train and  evaluation  setting are reported in the later section.

\section{Results}
\label{results}
As outlined in section \ref{evaluation}, we compute F1 and EM scores for all pairs of fine-tune and evaluation settings on all datasets, averaged across checkpoints generated in parallel. Table \ref{tab:squad} provides the evaluation results on translated SQuAD dataset while table \ref{tab:mmqa} provides results on the MMQA dataset. The baseline row in both tables provides the benchmark scores as reported in the paper \cite{latestMQA}. We also publish our results on our proposed benchmark dataset, XQuAD, in table \ref{tab:xquad}. Due to the unavailability of the source of the current state-of-the-art model, we could not compare our model with theirs on this dataset.  In all the tables, \textit{with $Q_H-P_H$ Aug} refers to mono-lingual Hindi QA augmentation while \textit{with $ Q_E-P_H$ Aug} refers to cross-lingual QA augmentation.
 
\begin{table}[h!]
\caption{Results of evaluation on test set variants of translated SQuAD dataset (dataset \ref{squad_sec}) for each fine-tune setup, compared against benchmark results \cite{latestMQA}. All values are averaged across results on 10 checkpoints. The highest score in each evaluation category is highlighted for ease of comparison.} 
\begin{subtable}[t]{3.1in}
\subcaption{EM scores for all the experimental settings}
\begin{tabular}{l|c|c|c|c}
\hline
Model settings & $Q_E-P_E$   & $Q_E-P_H$  &  $Q_H-P_E$  &  $Q_H-P_H$              \\ \hline
Baseline         & 53.15  & 45.34  & 44.19  & 51.34  \\ 
Zero Shot       & \textbf{88.76 } & 30.08  & 28.11  & 31.23  \\ 
\text{with $Q_H-P_H$ Aug.}       & 81.76  & 42.36  & 49.07  & \textbf{51.60} \\ 
\text{with $Q_E-P_H$ Aug.}       & 79.14  & \textbf{52.93 } & \textbf{51.10 } & 50.67  \\ \hline
\end{tabular}

\label{tab:squad_F1}
\end{subtable}
\begin{subtable}[t]{3.1in}
\subcaption{F1 scores for all the experimental settings}
\begin{tabular}{l|c|c|c|c}
\hline
Model settings & $Q_E-P_E$   & $Q_E-P_H$  &  $Q_H-P_E$  &  $Q_H-P_H$             \\ \hline
Baseline         & 57.29 & 50.24 & 48.21 & 53.87 \\ 
Zero Shot       & \textbf{94.56} & 38.23 & 37.15 & 41.08 \\ 
\text{with $Q_H-P_H$ Aug.}       & 90.24 & 52.32 & 56.89 & \textbf{63.59} \\ 
\text{with $Q_E-P_H$ Aug.}       & 87.96 & \textbf{64.51} & \textbf{59.19} & 62.15 \\ \hline
\end{tabular}

\label{tab:squad_EM}

\end{subtable}

\label{tab:squad}
\end{table}

\begin{table}[h!]
\centering
\caption{Results of evaluation on test set variants of MMQA dataset (dataset \ref{mmqa_sec}) for each fine-tune setup, compared against benchmark results \cite{latestMQA}. All values are averaged across results on 10 checkpoints. The highest score in each evaluation category is highlighted for ease of comparison.}  
\begin{subtable}[t]{3.1in}
\subcaption{EM scores for all the experimental settings}
\begin{tabular}{l|c|c|c|c}
\hline
Model settings & $Q_E-P_E$   & $Q_E-P_H$  &  $Q_H-P_E$  &  $Q_H-P_H$              \\ \hline
Benchmark          & 44.78  & 34.68  & 33.41  & 41.46  \\ 
Zero Shot       & 53.04 & 37.23  & 34.69  & 34.11  \\ 
\text{with $Q_H-P_H$ Aug.}       & \textbf{55.64}  & 44.37  & 47.66  & \textbf{46.41} \\ 
\text{with $Q_E-P_H$ Aug.}       & 54.32  & \textbf{45.12 } & \textbf{47.71}  & 45.71  \\ \hline
\end{tabular}

\label{tab:mmqa_F1}
\end{subtable}
\begin{subtable}[t]{3.1in}
\subcaption{F1 scores for all the experimental settings}
\begin{tabular}{l|c|c|c|c}
\hline
Model settings & $Q_E-P_E$   & $Q_E-P_H$  &  $Q_H-P_E$  &  $Q_H-P_H$             \\ \hline
Benchmark          & 50.27 & 37.89 & 37.02 & 48.14 \\ 
Zero Shot       & 71.95 & 51.51 & 52.33 & 50.97 \\ 
\text{with $Q_H-P_H$ Aug.}       & \textbf{74.43} & 61.39 & \textbf{65.06} & \textbf{64.34} \\ 
\text{with $Q_E-P_H$ Aug.}       & 73.26 & \textbf{63.75} & 64.37 & 63.81 \\ \hline
\end{tabular}

\label{tab:mmqa_EM}

\end{subtable}

\label{tab:mmqa}
\end{table}

\begin{table}[h!]
   \centering
       \caption{Results of evaluation on test set variants of XQuAD dataset (dataset \ref{xquadsec}) for each fine-tune setup. All values are averaged across results on 10 checkpoints. The highest score in each evaluation category is highlighted for ease of comparison.} 
\begin{subtable}[t]{3.1in}
\subcaption{EM scores for all the experimental settings.}
\begin{tabular}{l|c|c|c|c}
\hline
Model settings & $Q_E-P_E$   & $Q_E-P_H$  &  $Q_H-P_E$  &  $Q_H-P_H$              \\ \hline
Zero Shot       & 66.97 & 18.49  & 22.27  & 30.92  \\
\text{with $Q_H-P_H$ Aug.}       & \textbf{67.05}  & 33.87  & 39.66  & \textbf{46.47} \\ 
\text{with $Q_E-P_H$ Aug.}       & 64.29  & \textbf{44.71} & \textbf{41.01 }& 45.63  \\ \hline
\end{tabular}
\label{tab:xquad_F1}
\end{subtable}

\begin{subtable}[t]{3.1in}
\subcaption{F1 scores for all the experimental settings.}
\begin{tabular}{l|c|c|c|c}
\hline
Model settings & $Q_E-P_E$   & $Q_E-P_H$  &  $Q_H-P_E$  &  $Q_H-P_H$             \\ \hline
Zero Shot       & 76.61 & 24.05 & 31.06 & 40.85 \\
\text{with $Q_H-P_H$ Aug.}       & \textbf{78.83} & 45.79 & 49.42 & \textbf{61.08 }\\
\text{with $Q_E-P_H$ Aug.}       & 76.51 & \textbf{57.31} &\textbf{ 51.04} & 59.80 \\ \hline
\end{tabular}

\label{tab:xquad_EM}

\end{subtable}
    \label{tab:xquad}
\end{table}

\subsection{Qualitative Result}
Table \ref{tab:squad_qualt} provides example predictions containing the context snippet (with gold answer \textbf{emphasized}), question and proposed model's predicted answer. Wherever possible, we try to use common examples and also provide the answer reported by the baseline model in their paper \cite{latestMQA}. The prediction of the proposed model comes from the best performing fine-tune setting on that dataset and evaluation variant. As an example, for SQuAD dataset and $Q_H-P_H$ variant, we use a sample prediction from monolingual augmentation setting as it gives the best results. Besides, we have made an effort to provide sample predictions on all datasets (mentioned along with the question) and all multilingual variants (mono- and cross-lingual). Our examples also cover a diversity of question types- e.g. Q1, Q2 and Q4 are factoid (what and who types), while  Q3, Q4 and Q6 are short-descriptive (why) type. Our examples also cover questions with exact match (Q1, Q2 and Q4), partial match (Q3) and mismatch (Q5, Q6).

\begin{table}

\caption{Qualitative results of the best performing fine-tune setups with each example containing the question, snippet (with emphasized gold answer), and the predicted answers by both our proposed model and the baseline.} 
\includegraphics[scale=0.57]{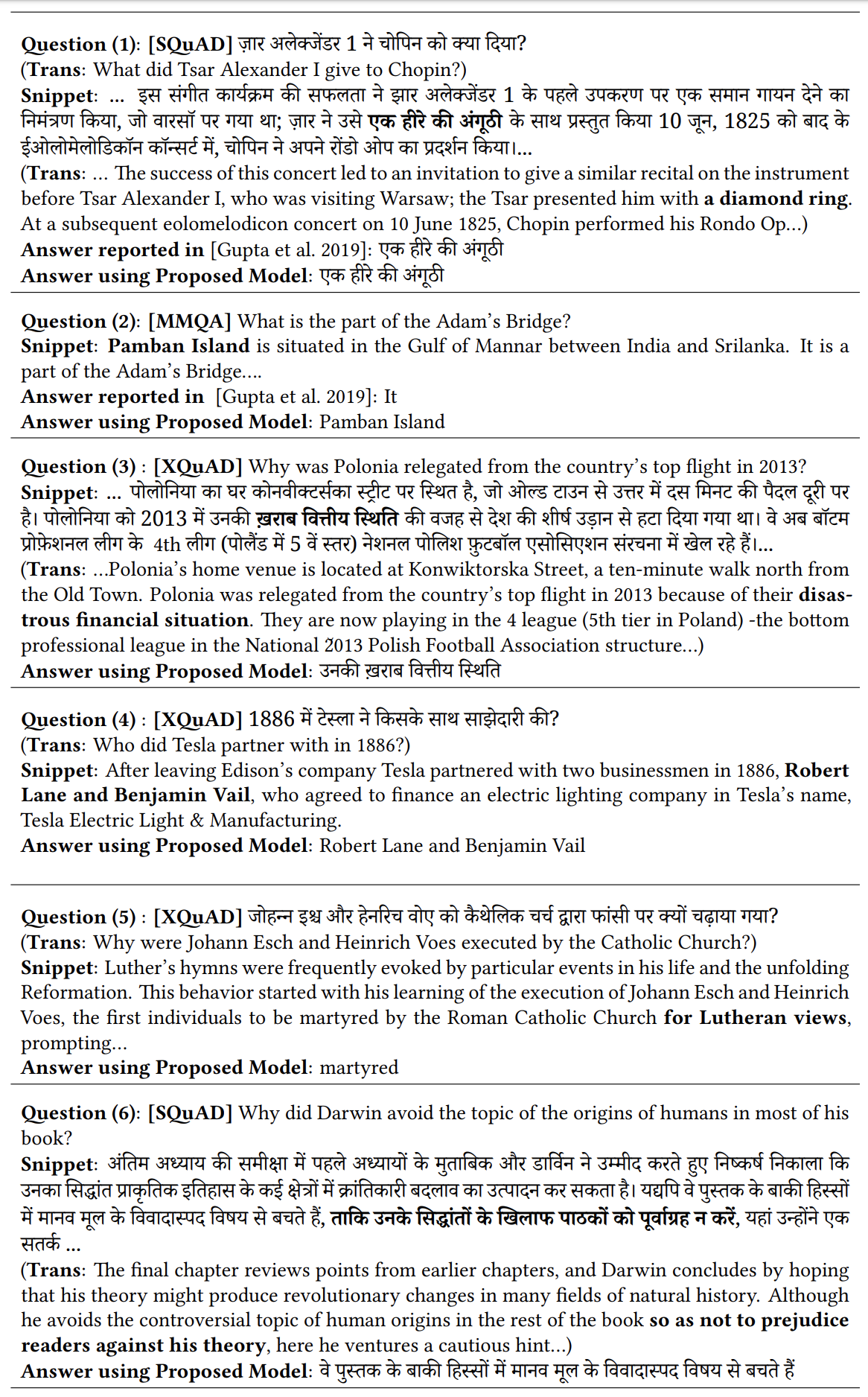}
 \label{tab:squad_qualt}
 \end{table}

\section{Discussion}
Following inferences can be drawn by analysing the quantitative and qualitative results published in section \ref{results}:
\begin{enumerate}
    \item \textit{Augmented m-BERT beats the baseline sequential model on all datasets} (SQuAD and MMQA). Zero-shot at m-BERT alone can be insufficient for the task, as evident in SQuAD results on all Hindi based variants which are way below the baseline results on those evaluation settings. 
    \item \textit{Explicitly fine-tuning m-BERT on a multilingual setting improves performance in that setting, but mildly degrades on the others}. This is  congruent with our motivation behind the fine-tune settings defined in section \ref{finetune_settings}. Therefore, monolingual Hindi QA augmentation improves performance on $Q_H-P_H$ while cross-lingual augmentation improves on cross-lingual evaluations on all datasets. Cross-lingual augmentation marginally degrades the mono-lingual results on both English and Hindi.  
    \item \textit{The size of training corpus used for fine-tuning impacts the final results.} This is corroborated by the fact that the net improvement brought by mono- or cross-lingual augmentations over their previous state results are way less compared to the improvement brought by SQuAD fine-tuning in zero-shot for $Q_E-P_E$, since SQuAD with 100K instances has 10 times more instances than training set for augmentations (10.5K). This also implies that m-BERT has potential for better results on Hindi with a larger training set.
    \item \textit{The choice of the cross-lingual variant to augment with does not impact our motive}: Although we choose $Q_E-P_H$ to fine-tune in cross-lingual augmentation setup, both the cross-lingual results ($Q_E-P_H$ and $Q_H-P_E$) are improved instead of reduction observed in other variants. It does, however, impact the net improvement in results. Considering that our prime goal was to show the potential of m-BERT, either of the variants hold good. 
    \item \textit{M-BERT based MMC performs well in cases of anaphora and cataphora resolution}. In Q2, the model is able to identify the antecedent ``Pumban Island" over ``It", while the baseline model gets stuck with the local reference, probably due to its sequential structure. In Q4, the model prefers to answer subsequent reference ``Robert Lane and Benjamin Vail" over the phrase ``two businessmen".
    \item Qualitatively, \textit{performance on factoid questions is much better than that of descriptive/reasoning questions}. Of the six questions, all factoid questions have exact match while only one descriptive question matches partially. This may be due to training set being skewed towards factoid type questions. An analysis of the translated SQuAD set reveals that descriptive questions constitute only 10\% (1.3K of the 10.5K) of the entire dataset. SQuAD, its parent set, contains only 11.9\% of such questions \cite{squad}.
    \item An \textit{error analysis} of the descriptive questions shows that although the results are incorrect, the \textit{model is able to understand the context of the question and provides a phrase that paraphrases the question}. While Q3 prediction is conceptually correct, ``martyred" for ``execution" in Q5 and literally restating the question in translation in Q6 show that model is able to develop a multilingual understanding but falls short in reasoning for the question. Since training set lacks such contextual reasoning questions, it would be a good direction to research whether m-BERT is capable of tackling such questions, given enough fine-tuning. 
    
\end{enumerate}
\section{Conclusion}

\par Our experiments with m-BERT for Multilingual Machine Comprehension task showed that m-BERT, augmented with English and Hindi multilingual QA (fine-tuned in order), significantly outperforms the current state-of-the-art sequential model \cite{latestMQA} for English and Hindi. We also adapted the recently released evaluation dataseet, XQuAD, for MQA by extending it with cross-lingual English-Hindi variants. Being a widely accepted and human translated dataset, we propose this extended version of XQuAD as evaluation benchmark and publish our results on this dataset as baseline for future research in this domain. Our analysis of this BERT variant may be beneficial for study on cross-lingual information retrieval involving low-resource languages like Hindi. 
\par Our research has two open-ended directions which can be pursued further. First, our analysis revealed that larger multilingual training set can improve results, thus indicating a scope for improvement in  multilingual comprehension of m-BERT for Hindi QA. Since no such large scale multilingual dataset exists for English and Hindi, future efforts can be made to extend XQuAD for training as well, by incorporating the entire SQuAD corpus. SQuAD, however, is severely limited in its resources for descriptive (non-factoid) questions. Therefore, another important research direction can be to evaluate m-BERT on a dataset which has sufficient multilingual descriptive questions to fine-tune on. 

\par We hope that our attempts towards setting a transfer learning based approach as the new baseline and standardization of evaluation will help surge this otherwise dormant domain of English-Hindi MMC.

\bibliographystyle{ACM-Reference-Format}
\bibliography{eng-hindi-mmc}
\end{document}